\documentclass{article}

\usepackage[nonatbib,final]{ml4mols_2020}

\usepackage[utf8]{inputenc} 
\usepackage[T1]{fontenc}    
\usepackage{hyperref}       
\usepackage{url}            
\usepackage{booktabs}       
\usepackage{amsfonts}       
\usepackage{nicefrac}       
\usepackage{microtype}      
\usepackage{graphicx}
\usepackage{wrapfig}
\usepackage{cite}

\title{High throughput screening with machine learning}

\author{
   Oleksandr~Gurbych \\
  SoftServe Inc.,\\
  2d Sadova Str., 79021 Lviv, Ukraine \\
  \texttt{ogurb@softserveinc.com} \\
  \And
  Maksym~Druchok \\
  SoftServe Inc.,\\
  2d Sadova Str., 79021 Lviv, Ukraine \\
  Institute for Condensed Matter Physics \\
  1 Svientsitskii Str., 79011, Lviv, Ukraine \\
  \texttt{mdruc@softserveinc.com} \\
  \And
  Dzvenymyra~Yarish \\
  SoftServe Inc.,\\
  2d Sadova Str., 79021 Lviv, Ukraine \\
  \texttt{dyari@softserveinc.com} \\
  \And
  Sofiya~Garkot \\
  Ukrainian Catholic University \\
  17 Svientsitskii Str., 79011, Lviv, Ukraine \\
  \texttt{garkot@ucu.edu.ua} \\
}

\begin{document}

\maketitle

\begin{abstract}

This study assesses the efficiency of several popular machine learning approaches in the prediction of molecular binding affinity: CatBoost, Graph Attention Neural Network, and Bidirectional Encoder Representations from Transformers.
The models were trained to predict binding affinities in terms of inhibition constants $K_i$ for pairs of proteins and small organic molecules.
First two approaches use thoroughly selected physico-chemical features, while the third one is based on textual molecular representations -- it is one of the first attempts to apply Transformer-based predictors for the binding affinity.
We also discuss the visualization of attention layers within the Transformer approach in order to highlight the molecular sites responsible for interactions.
All approaches are free from atomic spatial coordinates thus avoiding bias from known structures and being able to generalize for compounds with unknown conformations.
The achieved accuracy for all suggested approaches prove their potential in high throughput screening.

\end{abstract}

\section{Introduction}

Initial stages of drug discovery require localization of a disease cause, understanding of the molecular mechanism, then suggesting and testing drug leads.
On a cellular level, a drug molecule binds to a target biomolecule and amends its function.
Once the disease target has been identified a list of drug candidates is drafted and screened for the target-candidate binding affinities.
Running \textit{in silico} high throughput screening helps shrink the initial pool of \textit{in vitro} tests by eliminating weakly scored candidates.
We observe the increasing amount of physico- and bio-chemical data as well as difficulties with massive data arrays processing, which boosts the development and application of ML models to this field. 

One of the screening techniques is the assessment of receptor-ligand binding affinities.
The receptor-ligand binding can be illustrated with a classic example of competitive inhibition when malonate blocks the activity of succinate dehydrogenase.
Succinate dehydrogenase is an enzyme that participates in a reaction chain that provides energy for cells.
Its particular function is to conduct dehydrogenation on succinate molecules.
However, if the active site of dehydrogenase is occupied by slightly different malonate (one CH$_2$ group is missing) no catalytic reaction occurs.
Dehydrogenase is the receptor, whereas succinate and malonate are ligands competing for the active site of the receptor.
The ligand with stronger attraction (binding affinity) wins the competition deciding whether the reaction chain is activated or blocked.

There is a number of studies applying ML techniques to the affinity prediction problem~\cite{King1995,Jorissen2005,Yugandhar2014,Jimenez2018,Ozturk2018,LLi2019,YLi2019,Kwon2020} and the methods vary from supporting vector machine, decision trees, random forest to deep neural networks.
Of course, the list is not complete, it rather references representative papers in the field.
It is difficult to compare them directly since they use different metrics, affinity characteristics, and datasets.
In particular, the early researches started with only hundreds of affinity samples, while more recent studies advanced with growing availability of data.

In this contribution we test three machine learning algorithms -- CatBoost~\cite{Dorogush2018}, Graph Attention Neural Network, and Bidirectional Encoder Representations from Transformers~~\cite{Devlin2018,Devlin2019}.
All these techniques are used to predict the binding affinity expressed as inhibitory constant $K_i$.
We show that the considered approaches can be used for high throughput screening to save the time and resources on drug discovery stage by prioritizing more promising candidates for further development. 

\section{Materials and methods}

The dataset used in this study is compiled out of four databases -- BindingDB~\cite{Chen2001}, PDBbind~\cite{Wang2004}, Binding MOAD~\cite{Liegi2005}, BioLIP~\cite{Yang2013}
A plain concatenation of datasets results in a number of duplicated protein-ligand pairs, whereas the inhibition constants $K_i$ need to be inspected for nonphysical extreme concentrations.
The inhibition constants are ligand concentrations required to produce half maximum inhibition; they are expressed in nanomoles.
In concentration measurements the error increases proportionally to the concentration itself, therefore, it is convenient to work with decimal logarithm of inhibition constants $\log_{10} K_i$.
Such a conversion helps balance the error contributions at different concentration ranges.
After the whole data pre-processing the collected dataset consists of $\approx$350k
unique receptor-ligand pairs with corresponding values of $\log_{10}K_i$.

The dataset consists of receptors (target proteins) represented in FASTA format~\cite{Lipman1985,Pearson1988} and the ligands (inhibitors) -- in SMILES format~\cite{Weininger1988,Weininger1989}. 
Both formats are one-line sequence notations and do not use spatial coordinates; it was our essential requirement so that the trained models can make predictions on novel receptors and ligands with unknown conformations.
Otherwise, if coordinates are known one can develop coordinate-based predictors, like $K_{\rm DEEP}$ approach~\cite{Jimenez2018} splitting space into 3D grid and assigning each elementary cube a set of corresponding descriptors.

\begin{wrapfigure}{l}{0.40\textwidth}
\begin{center}
\includegraphics[clip=true,width=0.39\columnwidth]{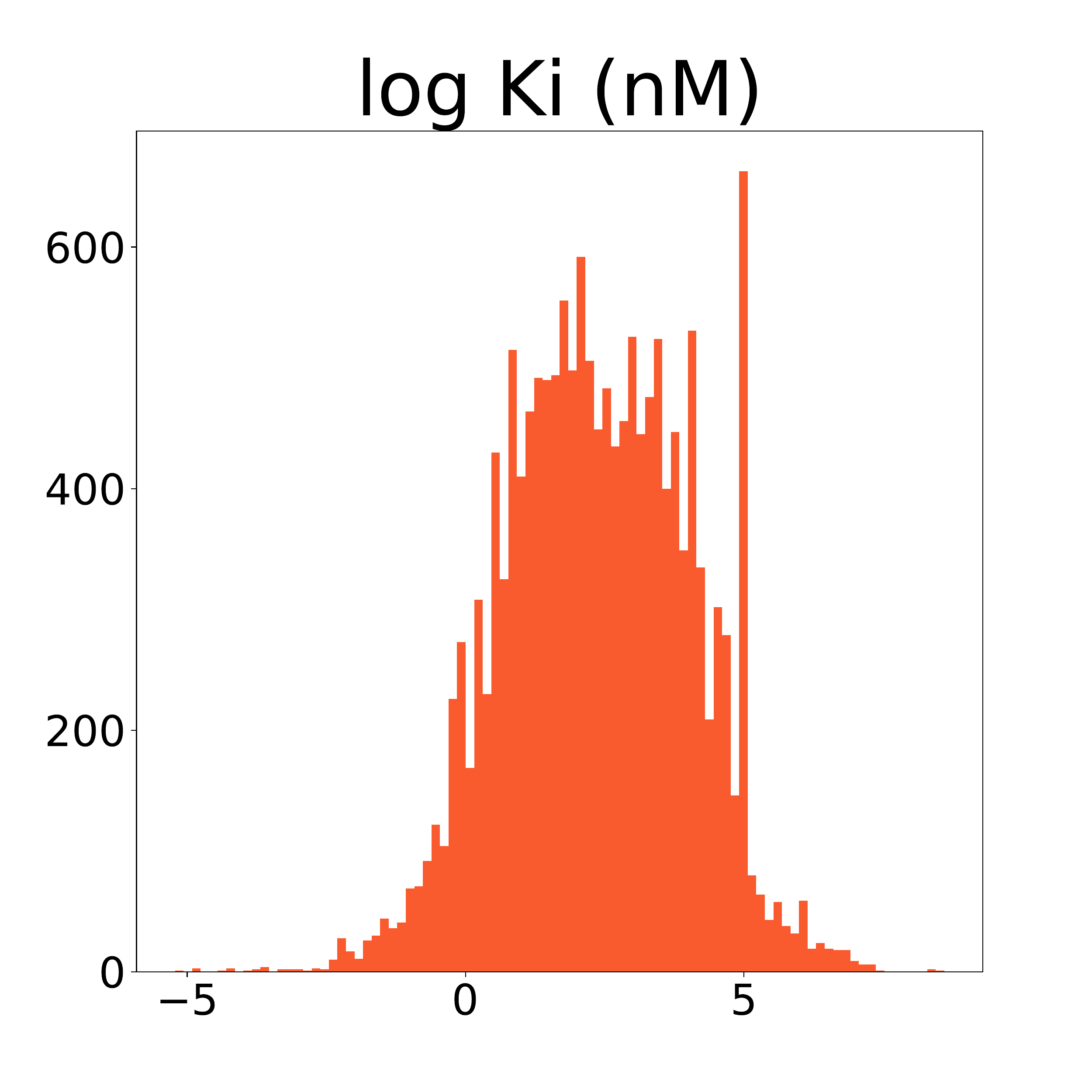}
\caption{Histogram for binding affinities as decimal logarithm of inhibition constants $\log_{10}K_i$. The bare $K_i$ constants are expressed in nanomoles.}
\label{fig:ki}
\end{center}
\end{wrapfigure}
In Fig.~\ref{fig:ki} we show the distribution of collected $\log_{10}K_i$ values.
It has symmetric shape with a slight imbalance for the regions of strong and moderate inhibitors ($\log_{10}K_i < 4$, i.e. $K_i < 10\mu$M), while inactive compounds (with $K_i$'s above 10$\mu$M) are less populated.
One of possible ways to balance the dataset is to generate augmentations:
most popular augmentations over SMILES strings are based on conformation variations or different atom enumerations (see for example Refs.~\cite{Hemmerich2000,Bjerrum2017}).
However, the ligand representations utilized in our study are insensitive to these techniques, thus we decided to keep the dataset in the original form rather than synthetically generating new inhibitors without knowledge about their inhibition constants.

The first machine learning technique of choice is \textbf{CatBoost Regressor}~(CB), a gradient boosting algorithm on decision trees.
We examined various physico-chemical features to represent receptors and ligands.
As a result, we chose ECFP4 fingerprints for ligands, while receptors were characterized by a set of features based on different properties of amino acids. 
These properties include charge, van der Waals volume, solvent accessible area, buried area, flexibility, folding index, hydrophobicity, isoelectric point, polarity, refractivity, aliphatic index, numbers of H-bond donors and acceptors, and number of disulfides.
All features for each receptor-ligand pair were combined into one vector and regressed towards corresponding $\log_{10} K_i$ value.

The second technique is \textbf{Graph Attention Neural Network} (GANN).
We designed the architecture of this model to take two inputs: featurized graphs for receptor and ligand.
The ligand graphs consist of nodes (all heavy atoms) and edges (all intramolecular bonds).
The ligand node and edge features encode atom and bond types, charges, number of electrons, hybridization, aromaticity, etc.
The receptor graphs contain only nodes (amino acids along the protein chain), since all peptide bonds uniting amino acids are identical.
The graphs for ligands were crafted with the help of standard featurizers, coming with DGL-LifeSci package (https://lifesci.dgl.ai/), whereas for the receptor graphs we developed a custom featurizer.
The receptor node features partially resemble the features used in the CB approach and include charge, flexibility, number of H-bond donors and acceptors, hydrophobicity, solvent accessible area, molecular weight, polarity, van der Waals volume, etc.
The graphs are fed into two separate ``arms''.
The receptor arm is based on Graph Attention network~\cite{Velickovic2017} from DGL-LifeSci.
In the SMILES arm we utilized the AttentiveFPGNN realization of Attentive FP network~\cite{Xiong2019} from the DGL-LifeSci package as well.
Outputs of these arms are then concatenated and regressed with a multilayer perceptron to predict a value of $\log_{10} K_i$.

In contrast to the two above approaches purely dealing with physico-chemical properties, \textbf{Bidirectional Encoder Representations from Transformers} (BERT) treats receptor and ligand sequences as text strings and constructs representations by jointly conditioning on both left and right context in all layers, which is crucial to fully capture the intricate inter-dependencies in the chemical structures. 
Our model is built of two pre-trained BERT heads (for FASTA and SMILES) and a multilayer perceptron on top of them.
BERT architecture is the same for two inputs: 6 layers, 12 self-attention heads, hidden size 768. 
FASTA BERT is pre-trained with the masked language modelling objective on the subset of UniProtKB protein sequences database, Swiss-Prot consisting of 506k entries~\cite{Bairoch2000TheSP}. 
Pre-trained SMILES BERT weights were taken from Ref.~\cite{Wolf2019}, optimized on 155k sequences from publicly available PubChem dataset~\cite{Kim2019PubChem2U}.
We concatenate the averaged final hidden states from FASTA BERT and SMILES BERT, and use it as an input to the perceptron, which then outputs $\log_{10}K_i$ value.
To our best knowledge, it is one of the first attempts to utilize Transformer-like models as sole predictors of the binding affinity.
Worth mentioning here a recent paper by Morris~\textit{et al.}~\cite{Morris2020} adopting Transformer approach for the affinity prediction, however, their setup is limited to single receptor task, thus embeddings are learned for ligand SMILES only.

\section{Results}

The dataset was randomly split into train, validation, and test subsets with the 80:10:10 ratios.
No specific pre-selection by receptor or ligand families was done.
The train and validation parts were used to train the ML models and monitor their accuracy during the train routine, whereas with the test subset we evaluated the final models.
Such a unified data split was intended to evaluate all three approaches on the same footing.
The best achieved Mean Absolute Error (MAE) scores (in $\log_{10}K_i$ units) on the test subset are CB — 0.60, GANN — 0.64, BERT — 0.51.
For example, the MAE of $\pm$0.6 for $\log_{10}K_i$ corresponds to a factor 4 (multiplier or divider) to the bare $K_i$ value, which is only a bit weaker accuracy than experimentally acceptable factor of $\le$2.
Worth noting that our values of MAE are lower than 0.71$\div$1.03 values obtained in Ref.~\cite{YLi2019} on PDBbind and Astex datatsets~\cite{Wang2004,Hartshorn2007}.

\begin{wrapfigure}{l}{0.42\textwidth}
\begin{center}
\includegraphics[clip=true,width=0.41\columnwidth]{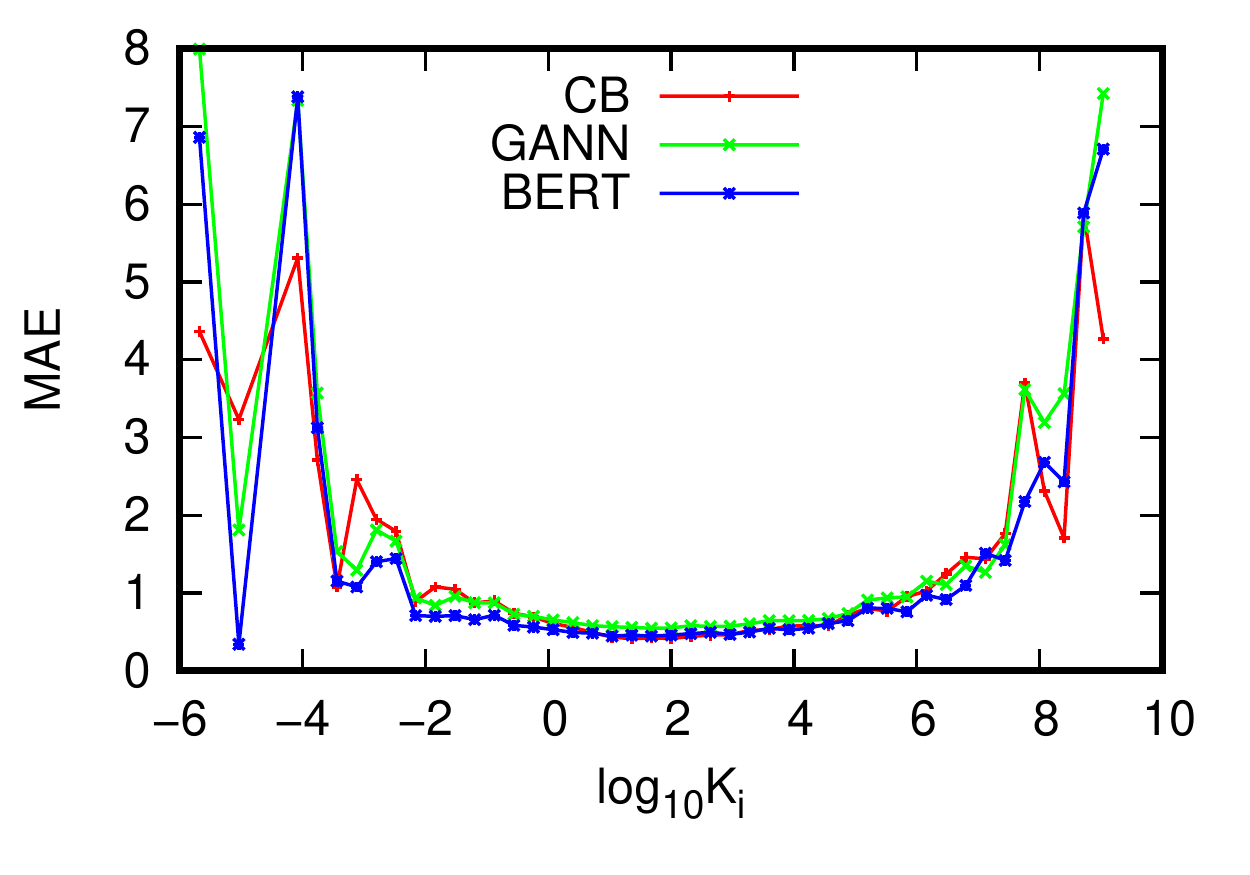}
\caption{Mean absolute errors as a function of experimental inhibition constants.}
\label{fig:mae}
\end{center}
\end{wrapfigure}
We plotted bin-averaged MAE as a function of actual $\log_{10}K_i$ for the three methods of choice (see Fig.~\ref{fig:mae}).
One can see high MAE peaks at low and high $\log_{10}K_i$, with a broad well inbetween.
We attribute these peaks to a low number of data points -- the ML models lack the examples to properly learn extreme cases (see Fig.~\ref{fig:ki}).

In a simpler case of a binary classification ``inhibitor vs. non-inhibitor'' (either $\log_{10}K_i$ is less or greater than 4) even high MAE values for low $\log_{10}K_i$ (left band of peaks in Fig.~\ref{fig:mae}) still provide a decent chance that compound will be correctly classified as inhibitor.
These are the high MAE values for high $\log_{10}K_i$ (peaks on the right) that pose a risk of misclassified type, since the boundary of $\log_{10}K_i = 4$ is closer to this weakly reproduced region.
In particular, the inhibitor vs. non-inhibitor classification yields accuracy of 94\% for CB and BERT, while GANN provides 93\%.
These values are comparable with accuracy of 95\% reported in a study by Li~\textit{et al.}~\cite{LLi2019} utilizing Bayesian additive regression trees.

\begin{figure}
\begin{center}
\includegraphics[clip=true,width=0.4\columnwidth]{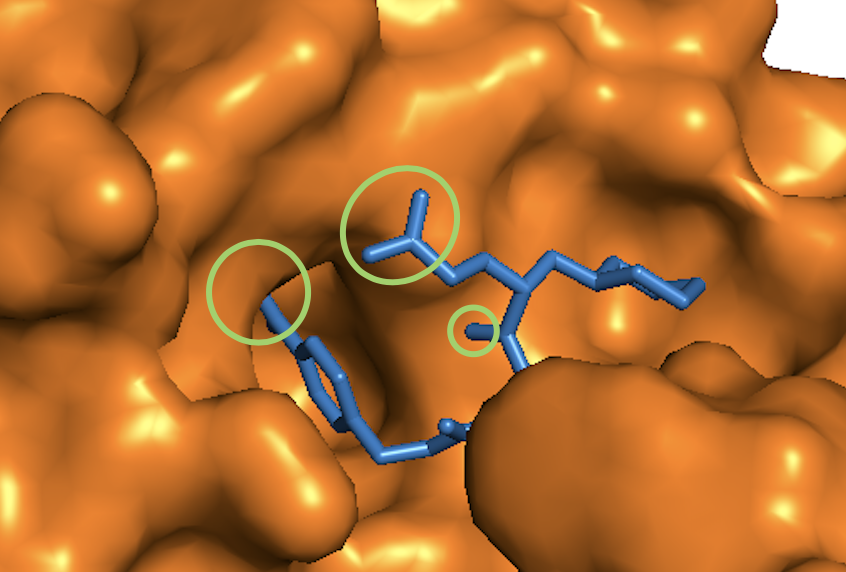}
\includegraphics[clip=true,width=0.4\columnwidth]{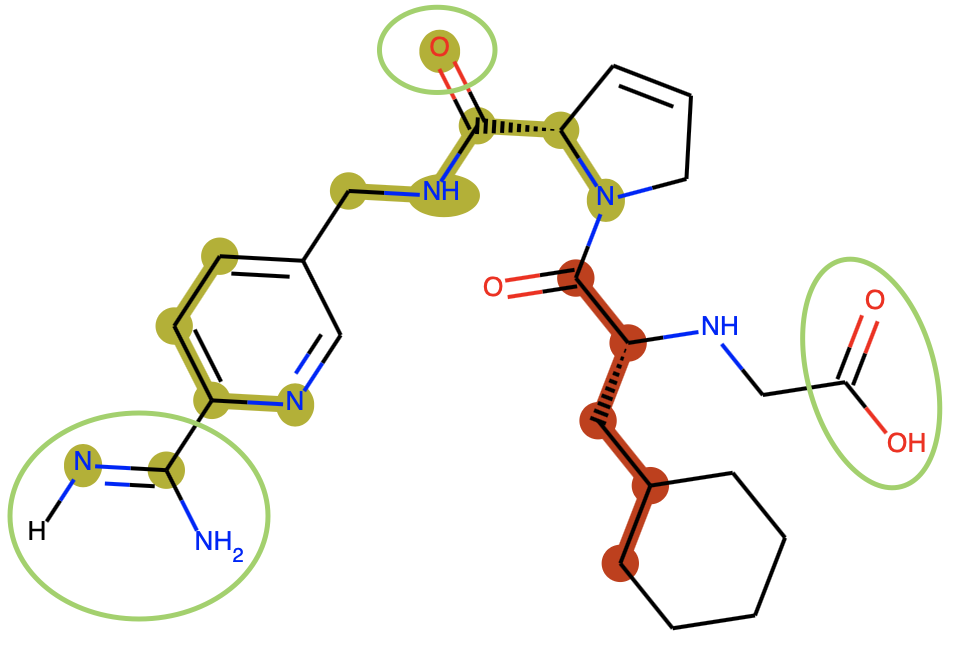}
\caption{Left -- fragment of human thrombin (orange surface) and a ligand  (blue sticks) occupying its active site with binding atoms (green ovals) defined experimentally.
Right -- 2D projection of the ligand with binding atoms highlighted by attention.
Green and red spots mark atoms with attention scores $<$5\% and $\ge$5\%.
The plots are made with PyMol (https://pymol.org) and RDKit (https://www.rdkit.org/) packages.}
\label{fig:bert_vis}
\end{center}
\end{figure}
We also visualized the attention scores ascribed by BERT model to ligands.
In general, the attention scores prioritize elements most relevant to a target task, thus comparing the scores for different ligand fragments we expected to highlight the ones responsible for binding.
However, we noticed that attention focus is often spread across ligand molecules instead of particular sites.
One of such examples is visualized in Fig.~\ref{fig:bert_vis} with a complex of human thrombin and a ligand taken from the Protein Data Bank (http://www.rcsb.org/structure/2ANM).
This ligand is known to be a strong inhibitor to the human thrombin.
The left snapshot shows an active site of thrombin molecule (depicted as an orange surface) occupied by the ligand (blue sticks), the green ovals show the ligand atoms interacting with the receptor (according to the Protein Data Bank).
The right plot shows a planar structure of the ligand: again, the green ovals denote the interacting atoms, while red and yellow spots indicate molecular fragments assumed by attention to contribute most to the prediction of affinity of the ligand to thrombin.
One can see that the attention is not only focused on the atoms in ovals.
It might showcase that binding is actually dependent on a full topology of a ligand rather than some particular sites.

\section{Conclusions}

This study is aimed to compare three popular machine learning methods in prediction of molecular binding affinity.
The following methods were tested: CatBoost, Graph Attention Neural Network, and Bidirectional Encoder Representations from Transformers.
For development of the first two we used physico-chemical features in representations of molecules, whereas the third one operated with FASTA and SMILES notations as text sequences.

The above approaches were tested on the same dataset with fixed split onto same train/validation/test subsets in order to maintain equal evaluation conditions.
The achieved accuracy for the three considered approaches shows a fair level of prediction, which is only slightly weaker than the acceptable experimental accuracy.
Besides the single-number metric Mean Absolute Error we also inspect how the deviations behave along the data range of affinities and map the regions of stronger and weaker confidence.
All the considered techniques prove to be lightweight and efficient for high throughput screening.

We also examine and visualize attention scores for ligands within the Transformer approach.
We expected them to indicate molecular sites responsible for interactions with proteins.
However, the attention often highlights broader parts of molecules, which might be an indication that binding mechanisms involve larger structural fragments.

Though the class of proteins is wide \textit{per se}, we believe that the considered approaches can be extended to a broader range of biomolecules, making them more universal.

\bibliographystyle{unsrt}
\bibliography{refs}

\end{document}